\documentclass[letterpaper]{article} % DO NOT CHANGE THIS

\usepackage[utf8]{inputenc} % Allow UTF-8 Unicode input
\DeclareUnicodeCharacter{FF0C}{,} % Map full-width comma to normal comma
\usepackage{aaai2026}  % DO NOT CHANGE THIS
\usepackage{times}  % DO NOT CHANGE THIS
\usepackage{helvet}  % DO NOT CHANGE THIS
\usepackage{courier}  % DO NOT CHANGE THIS
\usepackage[hyphens]{url}  % DO NOT CHANGE THIS
\usepackage{graphicx} % DO NOT CHANGE THIS
\usepackage{natbib}  % DO NOT CHANGE THIS AND DO NOT ADD ANY OPTIONS TO IT
\usepackage{caption} % DO NOT CHANGE THIS AND DO NOT ADD ANY OPTIONS TO IT
\usepackage{siunitx} % For S column type in tables

\usepackage{booktabs} % Added to support \toprule, \midrule, \bottomrule

\usepackage{algorithm}
\usepackage{algorithmic}

\usepackage{newfloat}
\usepackage{listings}
\DeclareCaptionStyle{ruled}{labelfont=normalfont,labelsep=colon,strut=off} % DO NOT CHANGE THIS
\floatstyle{ruled}
\newfloat{listing}{tb}{lst}{}
\floatname{listing}{Listing}
\title{EIDSeg: A Pixel-Level Semantic Segmentation Dataset for Post-Earthquake Damage Assessment from Social Media Images}
\author{
 Huili Huang\textsuperscript{\rm 1},
Chengeng Liu\textsuperscript{\rm 2},
Danrong Zhang\textsuperscript{\rm 1},
Shail Patel\textsuperscript{\rm 3},
Anastasiya Masalava\textsuperscript{\rm 3},
Sagar Sadak\textsuperscript{\rm 3} \\
Parisa Babolhavaeji\textsuperscript{\rm 3},
Weihong Low\textsuperscript{\rm 3},
Max Mahdi Roozbahani\textsuperscript{\rm 4},
J.~David Frost\textsuperscript{\rm 2}
}
\affiliations{
    \textsuperscript{\rm 1}School of Computational Science and Engineering, 
Georgia Institute of Technology, Atlanta, USA\\
\textsuperscript{\rm 2} School of Civil and Environmental Engineering, 
Georgia Institute of Technology, Atlanta, USA\\
\textsuperscript{\rm 3} School of Computer Science, 
Georgia Institute of Technology, Atlanta,USA  \\
\textsuperscript{\rm 4} School of Computing Instruction, Georgia Institute of Technology, Atlanta,USA\\

    \{hhaung413,cliu848\}@gatech.edu,Danrongzhang@microsoft.com,\\
\{spatel776, amasalava3, ssadak3, pbabolhavaeji3, wlow7,mahdir,jf36\}@gatech.edu
}

\usepackage{bibentry}
\DeclareUnicodeCharacter{202F}{\,}
\begin{document}

\maketitle

\begin{abstract}

Rapid post‑earthquake damage assessment is crucial for rescue and resource planning. Still, existing remote sensing methods depend on costly aerial images, expert labeling, and produce only binary damage maps for early-stage evaluation. Although ground-level images from social networks provide a valuable source to fill this gap, a large pixel-level annotated dataset for this task is still unavailable. We introduce \textbf{EIDSeg}, the first large-scale semantic segmentation dataset specifically for post-earthquake social media imagery. The dataset comprises 3,266 images from nine major earthquakes (2008–2023), annotated across five classes of infrastructure damage. Undamaged Building, Damaged Building, Destroyed Building, Undamaged Road, and Damaged Road. We propose a practical three-phase cross-disciplinary annotation protocol with labeling guidelines that enables consistent segmentation by non-expert annotators, achieving over 70\% inter-annotator agreement. We benchmark several state-of-the-art segmentation models, identifying Encoder-only Mask Transformer (EoMT) as the top-performing method with a Mean Intersection over Union (mIoU) of 80.8\%. By unlocking social networks' rich, ground-level perspective, our work paves the way for a faster, finer-grained damage assessment in the post-earthquake scenario.

\end{abstract}

\begin{links}
    \link{Code and Datasets}{https://github.com/HUILIHUANG413/EIDSeg}
    %\link{Datasets}{https://aaai.org/example/datasets}
    \link{Extended version}{https://arxiv.org/abs/2511.06456}
\end{links}

\section{Introduction}
Earthquakes cause an estimated \$100 billion in damage globally each year, posing a persistent challenge to the resilience of infrastructure and the response to disasters~\cite{GEM2023seismic}. Rapid post-disaster analysis and response are the keys to rescue efforts, resource deployment, and reducing economic loss in the immediate aftermath of a major earthquake. However, damage assessment efforts often face a “cold start” problem: getting the analysis off the ground is delayed by challenges such as limited image availability, unpredictable weather conditions, and the need for labor intensive expert assessments, amongst others.

Conventional data collection is based on remote sensing imagery such as satellites and drones, combined with expert interpretation to delineate damaged areas and estimate severity~\cite{Gupta2019xBD,Huang2025Fidelity}. Although widely adopted, these approaches face several well-documented limitations. First, acquiring high-resolution imagery is expensive and is often obstructed by weather conditions like clouds, fog, or rain, particularly in the critical early stages of a disaster~\cite{Gupta2019xBD,freddi2021innovations,drones8110638}. Second, damage analysis based on expert annotation is labor-intensive and requires substantial domain knowledge, making the process both time-consuming and resource-demanding. Recent studies have proposed automated tools for early disaster analysis, such as NASA’s Damage Proxy Maps (DPM)~\cite{NASA2020ARIA_DPM} and Microsoft's Damage Model~\cite{Microsoft2023TurkeyEarthquake}, but they typically return only a binary label that indicates whether an area is damaged, without describing the category of infrastructure or severity level.

In contrast, mobile phones and social media platforms now provide an additional ground-level stream of photographs that are collected within minutes of an event. Social media platforms like X (formerly Twitter) expose street-scale damage invisible to aerial images. Despite this potential, social media imagery remains underutilized due to the lack of task-specific datasets with pixel-level annotations. While datasets such as EID~\cite{Huang2025Fidelity}, MEDIC~\cite{Alam2021MEDIC}, and DAD~\cite{DAD-Dataset} offer multiclass labels for post-disaster damage, they are limited to image-level classification. Zhang et al. proposed the Damage Semantic Segmentation (DSS) dataset~\cite{Zhang2025PixelsDamageSeverity}, but it only contains 607 post-earthquake social media images without benchmark results. As a result, there are currently no established standards for semantic segmentation of ground-level post-earthquake imagery~\cite{Zhu2020MSNet}.

Developing a pixel-level segmentation benchmark for post-earthquake social media imagery presents two primary challenges. First, images vary widely in aspect ratio, resolution, and camera angle. Although annotators can typically assign a damage label to the most prominent building in the foreground, thoroughly segmenting all visible infrastructures, including distant or partially obstructed ones, requires significant manual effort. Second, neither existing standards in civil engineering for ground-level images nor semantic segmentation guidelines for aerial imagery are well suited to label infrastructure damage in ground-level social media images. Specifically, frameworks such as EMS-98~\cite{Grunthal1998EMS98} and HAZUS~\cite{FEMA2020HAZUS4_2_SP3} evaluate structural damage based on construction type (e.g., masonry, reinforced concrete), which is often unrecognizable in user-generated content. Prior aerial image segmentation datasets such as EarthquakeNet(drone-based)~\cite{Jiang2024EarthquakeNet} and xBD(satellite-based)~\cite{Gupta2019xBD} offer semantic segmentation annotations; however, their aerial perspectives do not apply directly to ground-level perspectives~\cite{Huang2025Fidelity}. These limitations underscore the need for an annotation protocol and dataset specifically designed for social media imagery, which is both scalable and accessible to non-expert annotators working under high-pressure early-stage post-earthquake scenarios.

To fill this gap, we introduce \textbf{EIDSeg}, the first large-scale semantic segmentation dataset of post-earthquake social media images. We curate the dataset by filtering low-quality images from the EID dataset~\cite{Huang2025Fidelity} and integrating high-quality samples from the DSS dataset~\cite{Zhang2025PixelsDamageSeverity}. To ensure our guideline is clear and accessible, six non-expert annotators with no civil engineering backgrounds labeled the dataset under the supervision of domain experts in post-earthquake engineering and infrastructure assessment. A three-phase quality control process was applied to enhance label consistency and reduce subjectivity.

The final EIDSeg dataset contains \textbf{3,266} images from nine major earthquakes (2008–2023), labeled with five pixel-level infrastructure damage classes: Undamaged Building (UD\_Building), Damaged Building (D\_Building), Destroyed Building (Debris), Undamaged Road (UD\_Road), and Damaged Road (D\_Road). To the best of our knowledge, EIDSeg is the largest publicly available semantic segmentation benchmark for post-earthquake social media imagery.

The inter-annotator analysis shows that the annotators achieved agreement over 70\% in the final stage of the annotation, validating that the annotation guideline that we offer is clear and reproducible. We benchmark several state-of-the-art semantic segmentation models, including DeepLabV3+~\cite{Chen2018DeepLabV3Plus}, SegFormer~\cite{Xie2021SegFormer}, Mask2Former~\cite{Cheng2022Mask2Former}, BEiT~\cite{Bao2022BEiT}, OneFormer~\cite{Jain2023OneFormer}, and EoMT (Encoder-only Mask Transformer)~\cite{Kerssies2025EoMT}. Of these, the EoMT achieves the best performance, with an mIoU of 80.8\% and a pixel accuracy of 90.3\%.

In summary, our main contributions are as follows.

\begin{itemize}
\item \textbf{Dataset:} We present \textbf{EIDSeg}, a benchmark of 3,266 post-earthquake social media images annotated with five-class pixel-level damage masks for infrastructure damage assessment.
\item \textbf{Annotation guideline:} We design a three-phase cross-disciplinary annotation protocol that enables consistent labeling by non-expert annotators.
\item \textbf{Benchmark:} We provide baseline performance using state-of-the-art semantic segmentation models to facilitate future comparisons.
\end{itemize}

\section{Related Work}\label{sec: eq_dataset}
\subsection{Earthquake Infrastructure Damage Assessment Scale}

Civil engineering experts rely on drone images and field visits to post-earthquake sites to assess structural damage of infrastructures. A widely used standard, the European Macroseismic Scale (EMS-98)~\cite{Grunthal1998EMS98}, defines five damage levels for buildings. G1 (Negligible to slight), G2 (Moderate), G3 (Substantial to heavy), G4 (Very heavy), and G5 (Destruction). The manual includes illustrative sketches and example images to guide human assessors. However, EMS-98 is limited in scope; it focuses solely on reinforced concrete and masonry buildings, excluding other types of infrastructure, such as transportation networks. FEMA's HAZUS ~\cite{FEMA2020HAZUS4_2_SP3} framework provides a more comprehensive classification system. It outlines 16 structural damage standards across 36 building categories, covering materials like wood, steel, and concrete. HAZUS also extends its damage evaluation to transportation and utility systems, using four severity levels: slight, moderate, extensive, and complete.

Despite their comprehensiveness, both EMS-98 and HAZUS present challenges when applied to social media image analysis. Many of the required attributes, such as type of construction, load-bearing wall failure, roof integrity, and the presence of hairline cracks, are not consistently visible in the images. Moreover, non-experts often struggle to identify specific infrastructure types based solely on partial or external views. This makes a direct application of detailed engineering taxonomies impractical. %Adding to the issue, social networks and field images vary widely in viewpoint and scale, ranging from close-ups of flooded streets to aerial cityscapes, further complicating the consistent interpretation of damage.

\begin{table*}[!t]
\small
\centering
\setlength{\tabcolsep}{4pt}  % tighten column spacing
\begin{tabular}{l l c c l c c}
\toprule
\textbf{Dataset} & \textbf{Data Source} & \textbf{Year} & \textbf{Resolution} &
\textbf{Task} & \textbf{\# Post-EQ Images} & \textbf{\# Damaged Classes} \\
\midrule
xBD~\cite{Gupta2019xBD}          & Satellite                           & 2019 & 1024$\times$1024 & Instance Seg.\,/\,Cls.        & 192    & 4 \\

TUE-CD~\cite{Liu2024BuildingChangeDetectionTUECD}        & Satellite                           & 2024 & 256$\times$256   & Semantic Seg.          & 2338     & 2 \\

TE23D~\cite{Ekkazan2025TE23D}         & Satellite                           & 2025 & 256$\times$256   & Semantic Seg.         & 1,183  & 2 \\

EarthquakeNet~\cite{Jiang2024EarthquakeNet} & UAV                                 & 2024 & 5616$\times$3744 & Semantic Seg.          & 69     & 8 \\
MEDIC ~\cite{Alam2021MEDIC}        & Social Media                        & 2021 & Varies           & Multi-Task Cls.       & 15,095 & 3 \\
EID~\cite{Huang2025Fidelity}           & Social Media                        & 2025 & Varies           & Cls.                  & 13,513 & 4 \\
DSS~\cite{Zhang2025PixelsDamageSeverity}           & Social Media                        & 2025 & Varies           & Semantic Seg.          & 607    & 3 \\
\midrule
EIDSeg(Ours)       & Social Media                        & 2025 & Varies           & Semantic Seg.          & 3,266  & 5 \\
\bottomrule
\end{tabular}
\caption{Overview of publicly available earthquake-damage datasets.  ``Seg. = Segmentation", ``Cls." = classification,  ``Post-EQ" = post-earthquake. }
\label{tab:earthquake_datasets}
\end{table*}

\subsection{Earthquake Infrastructure Damage Dataset}

% Even though the large-scale datasets collect images from various disasters senarios can provide meaningful information in damage analysis, it may not be optimal for earthquake specific evaluation. For example, the characteristics of damage caused by earthquakes differ from those caused by floods\cite{te23d}.

Table \ref{tab:earthquake_datasets} shows an overview of the existing data for the assessment of post-earthquake damage. Most existing datasets for earthquake damage segmentation are either part of broader multi-disaster collections or limited to a single geographic event. For example, the widely used xBD dataset~\cite{Gupta2019xBD} covers 19 disaster events, comprising 22,068 satellite images and 850,736 annotated building polygons. However, it includes only 192 pre-event and 192 post-event satellite images from the 2017 Mexico City earthquake, which is insufficient for large-scale earthquake-specific analysis. More recent segmentation datasets, such as TUE-CD~\cite{Liu2024BuildingChangeDetectionTUECD} and TE23D~\cite{Ekkazan2025TE23D}, focus specifically on the 2023 Türkiye earthquakes but are geographically constrained, limiting their generalizability to other seismic regions.

Several datasets focus on the drone imagery. For instance, EarthquakeNet~\cite{Jiang2024EarthquakeNet} provides high-resolution UAV-based images from Baoxing County with detailed annotations but includes only 69 images. Other datasets, such as AIDER~\cite{Kyrkou2019AIDER} and AIDERv2~\cite{Shianios2023AIDERv2}, include earthquake-related classes such as \emph{collapsed buildings} to support the prediction of emergency response. However, these datasets lack damage severity labels, and the source events of the images are not specified, which limits their utility for systematic damage analysis.

For social media imagery, most studies have focused on image-level classification of damage severity~\cite{DAD-Dataset,Alam2017Image4ActOS,Nguyen2017AutomaticIF,Nia2017BuildingDA,Alam2021MEDIC,crisisdataset2020icwsm}. Huang et al. developed the Earthquake Infrastructure Damage (EID) dataset, offering detailed labeling guidelines tailored for non-expert annotators and achieving strong performance in disaster analysis tasks~\cite{Huang2025Fidelity}. For segmentation, Zhang et al. introduced the DSS dataset, which provides pixel-level segmentation masks tailored for post-earthquake damage assessment~\cite{Zhang2025PixelsDamageSeverity}. However, DSS lacks a formal benchmark and contains only 607 annotated images, limiting its generalization potential.

\section{Data Details}

\begin{figure*}[t]
    \centering
    \includegraphics[width=1.0\textwidth]{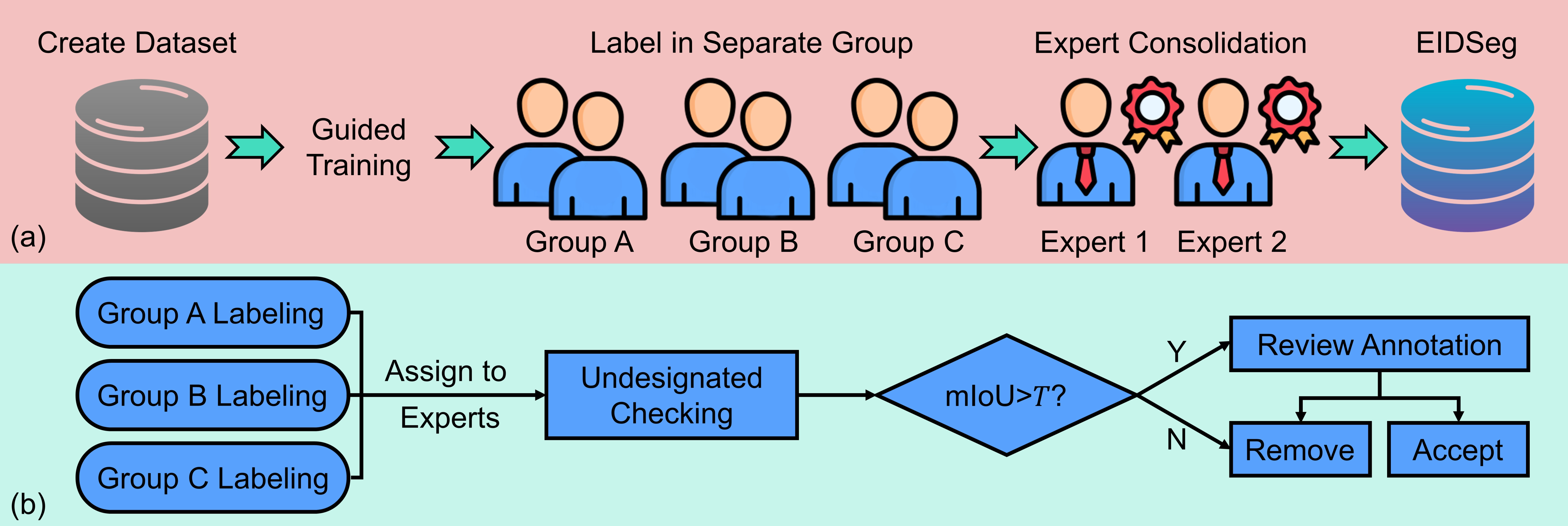}
    \caption{Annotation workflow of EIDSeg. (a): Dataset creation: images are collected and filtered from the EID and DSS datasets, then independently labeled by annotation groups (A–C). Experts evaluate label quality, and high-quality samples are curated into the final EIDSeg dataset.
    (b): Label evaluation: Annotations undergo experts review following an ``Undesignated Checking" step, which address uncertain labels provided by annotators. If inter-annotator agreement (mIoU) exceeds a threshold $T$, the image proceeds to expert adjudication; otherwise, it is discarded to ensure dataset integrity.}
    \label{fig:dataset_pipline}
\end{figure*}

\subsection{Data Collection and Cleaning}
EIDSeg integrates images from both the EID dataset~\cite{Huang2025Fidelity} and the DSS dataset~\cite{Zhang2025PixelsDamageSeverity}. It comprises images from nine major earthquakes that span 2008 to 2023, enhancing the generalizability of the dataset in diverse earthquake scenarios. 

The EID dataset compiles social media images from multiple sources, applying strict quality control and a well-documented annotation protocol. It covers five major events: the 2015 Nepal, 2015 Illapel, 2016 Ecuador, 2017 Mexico, and 2017 Iran–Iraq earthquakes. Each image is categorized into one of four classes: Irrelevant or Non-informative (DS0), No Damage (DS1), Mild Damage (DS2), and Severe Damage (DS3). Compared to earlier image-level datasets such as MEDIC~\cite{Alam2021MEDIC}, DAD~\cite{DAD-Dataset}, and CrisisBenchmark~\cite{crisisdataset2020icwsm}, EID provides finer class granularity and higher annotation quality. For EIDSeg, we include only DS1 to DS3 images, since DS0 images are mostly advertisements, maps, or logos that lack relevant infrastructure content.

The DSS dataset adds 607 images drawn from the 2008 Wenchuan, 2010 Haiti, 2015 Nepal, 2023 Morocco, and 2023 Türkiye earthquake, with additional examples retrieved through a Google search for ``earthquake damage". DSS assigns three segmentation labels: Undamaged Structure, Damaged Structure, and Debris. 

Labeling pixel-level masks for disaster damage is substantially more challenging than assigning image-level labels. Although annotators can often identify and categorize damage to prominent foreground structures (e.g., collapsed buildings or visible debris) as \textit{severe}, labeling distant or partially occluded infrastructure, such as roads covered in dust or background buildings, requires finer visual detail and greater annotation effort. To ensure reliable pixel-level annotations, we implemented an automated quality control filter that developed in consultation with domain experts to retain only high-quality, information-rich images from the raw dataset. Specifically, we excluded images that failed the following quality criteria:

\begin{itemize}
   \item \textbf{Sharpness.}  
         We followed standard practice in blur detection by computing the variance of the Laplacian operator on grayscale-converted images using OpenCV. Images with a variance below 50 were flagged as blurry.
   \item \textbf{Resolution.}  
          We enforced a minimum resolution of $512 \times 256$ pixels to eliminate low-resolution thumbnails and excessively cropped images. The distribution of image resolution are shown in Appendix C. 
\end{itemize}

Images failing either criterion were moved to a separate directory for manual review and excluded from the primary annotation set. This lightweight filtering process ensured that the initial dataset used for labeling maintained a consistent level of visual clarity necessary for accurate and consistent pixel-level labeling. As a result, 6,646 images were left for annotation.

\subsection{Damage Scale}
\begin{figure*}[htbp]
    \centering
    \includegraphics[width=1.0 \textwidth]{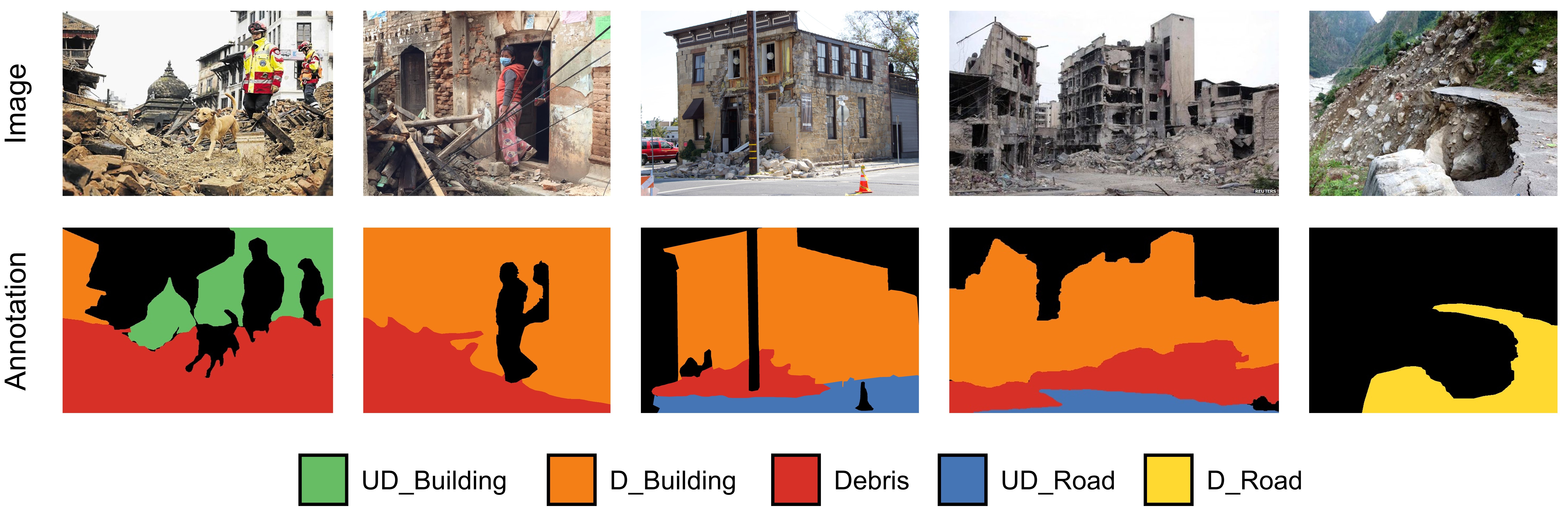} % Replace with your actual path and filename
    \caption{Damage Mask examples of EIDSeg.}
    \label{fig:eidseg_mask}
\end{figure*}

\begin{figure*}[t]
    \centering
    \includegraphics[width=0.85\textwidth]{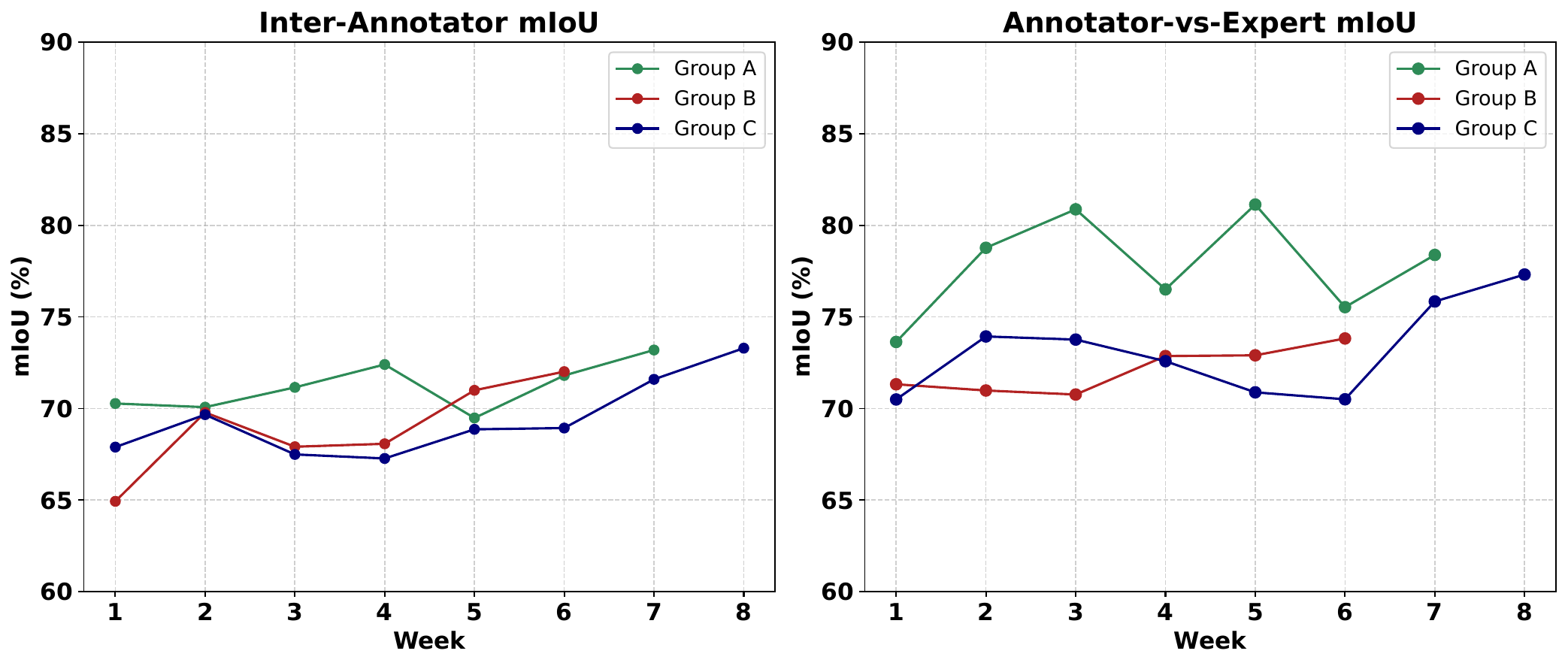}
    \caption{
        Weekly segmentation agreement across annotators. 
        \textbf{Left:} Inter-annotator mIoU (\%) among Groups A, B, and C. 
        \textbf{Right:} Average annotator–expert mIoU (\%) within each group over time.  Note that some annotators completed the assigned tasks earlier, which explains the different end dates in the figure (Group A in Week 8, and Group B in Weeks 7–8).
    }
    \label{fig:miou_agreement}
\end{figure*}

Referring to previous civil engineering manuals and remote sensing studies~\cite{Grunthal1998EMS98,FEMA2020HAZUS4_2_SP3,Jiang2024EarthquakeNet,Zhang2025PixelsDamageSeverity}, the EIDSeg dataset focuses on two common infrastructure types frequently observed in social media imagery: \textbf{buildings} and \textbf{roads}.

\textbf{Building structures} refer to enclosed spaces with walls and a roof used for living, working, or storage, such as houses, offices, warehouses, dams, power plants, and similar stationary structures. In post-disaster scenarios, damage may leave these structures to exposed frameworks or partial ruins. These remnants are still classified as building structures since they remain identifiable.

\textbf{Road structures} are defined as any paths or routes designated for transportation. This includes bridges, highways, streets, sidewalks, and bike paths, with emphasis on linear pathways.

Each instance of the infrastructure is annotated into one of five damage-level classes: Undamaged Building, Damaged Building, Destroyed Building, Undamaged Road and Damaged Road. The detailed definitions of these classes are as follows:

\begin{itemize}
    \item \textbf{Undamaged Building Structure (UD\_Building):} Building structures that show no visible signs of disaster-related damage. Note that small cracks may not be visible in social media images due to resolution limits, and such cases are still considered undamaged.
    
    \item \textbf{Damaged Building Structure (D\_Building):} Building structures that exhibit partial to significant damage that requires repair or restoration. These structures remain identifiable and distinguishable from debris.
    
    \item \textbf{Destroyed Building Structure (Debris):} Structures that are no longer identifiable due to complete collapse. In cases like pancake collapses, where floors have compressed into layered debris, the entire area is labeled as debris.
    
    \item \textbf{Undamaged Road Structure (UD\_Road):} Roads that appear to be free of visible cracks, potholes, or major surface disruptions. Minor imperfections, such as hairline cracks (e.g., less than 1/8 inch~\cite{FEMA2020HAZUS4_2_SP3}), are acceptable if they do not compromise structural integrity or user safety.
    
    \item \textbf{Damaged Road Structure (D\_Road):} Roads showing clear signs of damage, including large or extensive cracks, potholes, or surface deformations that affect functionality or safety and require repair.
    
     \item \textbf{Undesignated:} Structures that are visually identifiable as roads or buildings but lack sufficient visual clarity to determine their damage levels with confidence. This uncertainty may result from image blur, occlusion, or limited visibility (e.g., only a portion of the structure is captured in the image).
\end{itemize}

Note that the EIDSeg damage scale defines only two categories for roads (\textit{undamaged} and \textit{damaged}) to align with what is visually structurally significant in post-disaster imagery. As shown in the rightmost column of Figure \ref{fig:eidseg_mask}, roads generally do not undergo complete structural collapse as buildings do, which often results in clearly visible rubble or debris. Instead, road damage often appears as partial deterioration~\cite{Jiang2024EarthquakeNet}. Additionally, the presence of a ``debris-like" object on a roadway, such as collapsed structures or landslide material, does not necessarily imply that the road itself is damaged. In many cases, such rubbles originate from surrounding terrain (e.g., hillsides or mountains), rather than from the road infrastructure itself, and are therefore not included in our structural damage categories.

In addition to the structural damage classes, we introduce an \textit{Undesignated} class to reduce annotation ambiguity and ease the labeling burden in challenging cases. To maintain consistency and quality, all \textit{Undesignated} instances are reviewed by domain experts during each annotation round, allowing for a careful and conservative judgment. This class is excluded from the final EIDSeg dataset statistics. For all other annotation procedures, we follow the VOC2011 segmentation guidelines~\cite{Everingham2010VOC}.

\subsection{Annotation Process}

 Figure \ref{fig:dataset_pipline} illustrates the overall pipeline for the data annotation process. We generated the EIDSeg mask through a three-phase annotation protocol designed with expert quality control. Six non-expert annotators with no civil engineering backgrounds carried out the labeling under the supervision of two domain experts with expertise spanning both civil engineering and computer science.

\textbf{Phase 1: Guided training.}
For the first three weeks, all six annotators underwent training using a 300-image practice set that had been thoroughly labeled by the experts. Each annotator worked independently and their results were evaluated weekly. We computed the mIoU between the masks of each annotator and the expert reference masks to assess progress and identify potential misunderstandings of the labeling guideline. This process allowed us to monitor progress and uncover any misinterpretations of the labeling guidelines. Based on these evaluations, the experts also refined the guidelines to improve clarity. This iterative feedback loop helped minimize misunderstandings and ensured that annotators could accurately identify infrastructure types and assess the severity of the damage before moving onto the full data set.

\textbf{Phase 2: Production labeling with embedded gold images.}
Following the guided training phase, the six annotators were divided into three independent pairs (Groups A, B, and C) for full-scale labeling. Each week constituted one annotation round. In each round, the two annotators in a pair first labeled their assigned images independently. The experts then conducted review sessions with each group to address disagreements and clarify ambiguous cases. Annotators revised any identified errors, retained confident labels, and resubmitted the batch for final expert adjudication. The examples of annotator misalignment observed during Phase 2 are shown in Appendix A.

To ensure annotation quality, the experts prepared a curated set of 100 ``\textit{gold standard}" images. For each annotation round, they randomly selected 20–25 gold images and embedded them in the labeling batches assigned to each group. Two agreement metrics were tracked: (i) annotator-vs-expert mIoU (on gold standard images), and (ii) inter-annotator mIoU (on jointly labeled images). Annotators with expert agreement below 0.60 were flagged for retraining. Images with inter-annotator agreement less than 0.60 were excluded from the final dataset to avoid the subjectivity issue ~\cite{Huang2025Fidelity}. The $0.60$ threshold was selected based on the highest annotator-expert agreement observed during Phase 1 training. This ensured that only consistent annotations were carried forward to Phase 3.

\textbf{Phase 3: Expert consolidation.}
In the final phase, all images containing masks labeled as \textit{Undesignated} underwent the Undesignated Checking process, during which experts manually reviewed the uncertain labels and removed images with excessive ambiguity. For example, if a building labeled as \textit{Undesignated} remained indeterminate even after expert review, and the proportion of \textit{Undesignated} pixels exceeded 2\% of the image, the image was excluded from the dataset. Following this, experts conducted a comprehensive review of all remaining annotations, applying minor boundary refinements to improve precision and ensure consistency.

This multistage process ensures that each image in EIDSeg has been reviewed at least twice, first by trained annotators and then by expert reviewers, resulting in a high-quality, pixel-level segmentation benchmark with strong reliability.
\subsection{Targeted Dataset Statistics}
Figure~\ref{fig:eidseg_mask} presents several representative examples from the EIDSeg dataset, along with their corresponding color-coded annotation masks. Figure \ref{fig:miou_agreement} illustrates the average annotator-vs-expert mIoU and inter-annotator mIoU for each group over time during Phase 2. The agreement remained consistently high throughout the annotation process and exhibited an upward trend, indicating improved annotator consistency and better alignment with the expert-labeled ground truth over time.  By the final round, all groups achieved agreement above 70\%, a promising outcome for disaster assessment tasks. For comparison, the image-level EID dataset reached 86.6\% agreement in its final round~\cite{Huang2025Fidelity}, while EIDSeg demands more complex pixel-level labeling across finer-grained classes.

\begin{figure}[ht!]
    \centering
    \includegraphics[ width=0.95\linewidth]{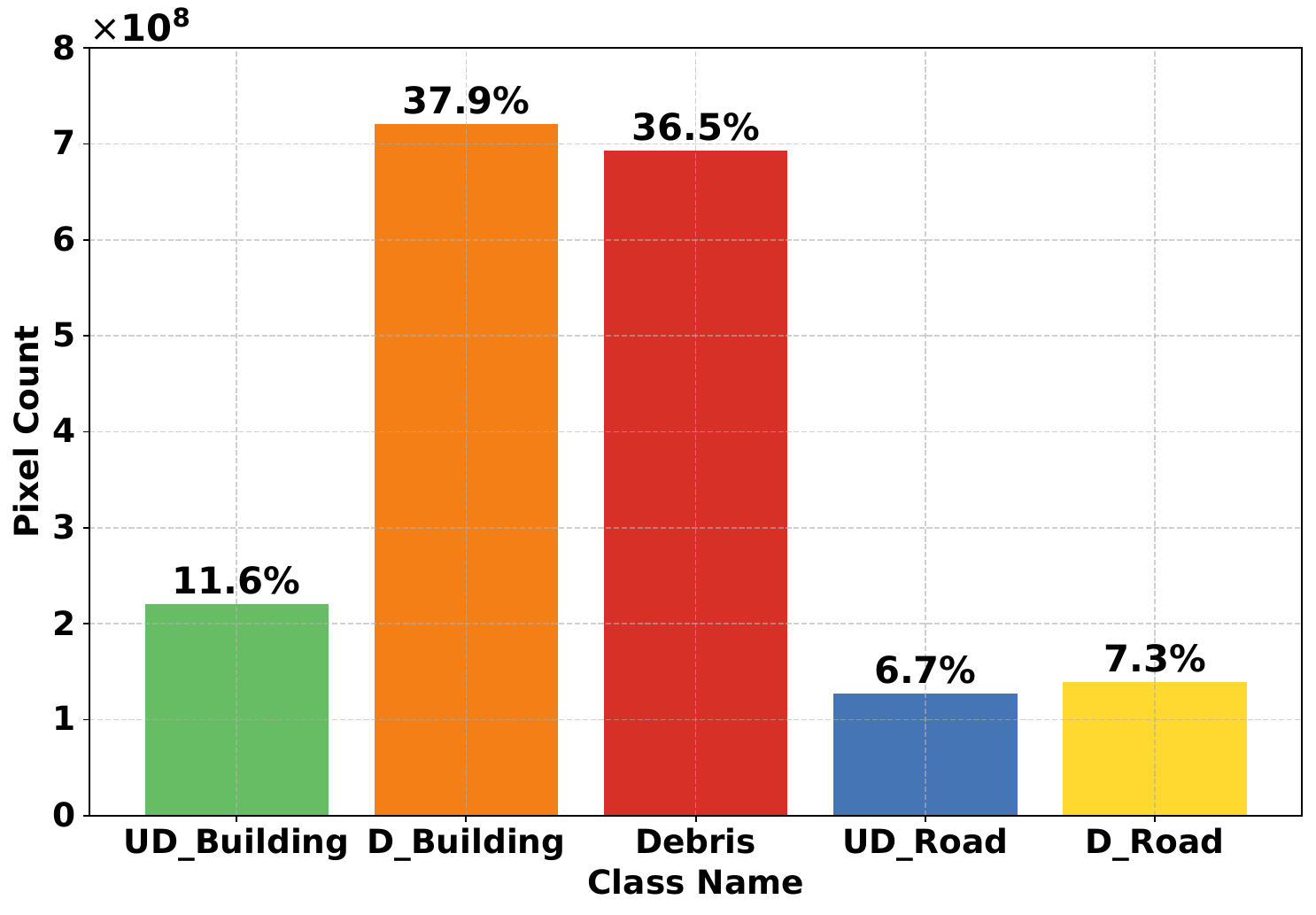}
    \caption{
        Pixel distribution across segmentation classes in the EIDSeg dataset.
        Each bar shows the total pixel count per class, along with the percentage it contributes to the full dataset.
        Class colors are chosen to reflect semantic segmentation label: \emph{Undamaged Building}, \emph{Damaged Building}, \emph{Debris}; and \emph{Undamaged Road}, \emph{Damaged Road}.
    }
    \label{fig:pixel_distribution}
\end{figure}

Figure~\ref{fig:pixel_distribution} displays the pixel distribution in EIDSeg. In particular, the dataset contains a higher proportion of pixels labeled damaged compared to undamaged. This trend reflects the underlying class distribution of the EID image corpus, which contains more disaster-affected scenes (52.1\%) than undamaged ones (26.86\%)~\cite{Huang2025Fidelity}. This suggests that social media users are more likely to capture and share images of visible damage than undisturbed scenes after an earthquake.

\section{Experiment}
\subsection{Experimental Setup}
\begin{table*}[htb]
  \centering
  \small
  \setlength{\tabcolsep}{5pt}
  \begin{tabular}{
      l                       % Model
      l                       % Backbone
      l                       % Pre‑train
      c                       % Input size
      S                       % mIoU
      S                       % FWIoU
      S                       % PA
      S                       % FLOPs
      S                       % Params
  }
    \toprule
    \textbf{Model} & \textbf{Backbone} & \textbf{Pre‑train} & \textbf{Input} &
    \multicolumn{1}{c}{\textbf{mIoU\,(\%)}} &
    \multicolumn{1}{c}{\textbf{FWIoU\,(\%)}} &
    \multicolumn{1}{c}{\textbf{PA\,(\%)}} &
    \multicolumn{1}{c}{\textbf{FLOPs\,(G)}} &
    \multicolumn{1}{c}{\textbf{Params\,(M)}}\\
    \midrule
    DeepLabV3~\cite{Chen2018DeepLabV3Plus}          & ResNet‑101 & Cityscapes & \(512^2\)  & 67.1 & 68.2 & 80.6 &  79.29  &  58.75 \\
    SegFormer~\cite{Xie2021SegFormer}          & MiT‑B5     & Cityscapes & \(512^2\)  & 74.4 & 75.2 & 86.9 & 110.16  &  84.60 \\
    Mask2Former~\cite{Cheng2022Mask2Former}        & Swin‑S     & Cityscapes & \(512^2\)  & 76.1 & 77.1 & 87.5 &  93.21  &  68.72 \\
    Mask2Former                & Swin‑L     & Cityscapes & \(512^2\)  & 77.4 & 78.4 & 88.7 & 250.54  & 215.45 \\
    BEiT~\cite{Bao2022BEiT}               & ViT‑B      & ADE20K     & \(640^2\)  & 77.7 & 78.1 & 88.8 & 890.06  & 163.26 \\
    BEiT                       & ViT‑L      & ADE20K     & \(640^2\)  & 78.7 & 79.6 & 89.5 & 1823.53 & 441.09 \\
    OneFormer~\cite{Jain2023OneFormer}          & Swin‑L     & Cityscapes & \(512^2\)  & 79.2 & 78.3 & 88.2 & 1042.14 & 218.77 \\
    EoMT~\cite{Kerssies2025EoMT}         & ViT‑L      & Cityscapes & \(1024^2\) & \bfseries 80.8 & \bfseries 80.9 & \bfseries 90.3 & 1341.85 & 319.02 \\
    \bottomrule
  \end{tabular}
  \caption{Semantic Segmentation Benchmark of \textit{EIDSeg}. Best scores are \textbf{bold}.}
  \label{tab:seg_results}
\end{table*}

\begin{table*}[htb]
\centering
\small
\setlength{\tabcolsep}{4pt}
\begin{tabular}{l c c c c c c}
\toprule
Model & UD\_Building & D\_Building & Debris & UD\_Road & D\_Road & mIoU (\%)\\
\midrule
DeepLabV3+ & 43.5 & 65.4 & 77.3 & 75.7 & 73.7 & 67.1\\
SegFormer         & 54.9 & 73.5 & 82.3 & 79.9 & 81.3 & 74.4\\
Mask2Former-S & 58.9 & 76.3 & 83.4 & 79.3 & 82.8 & 76.1\\
Mask2Former-L & 63.5 & 76.9 & 84.3 & 79.6 & 82.7 & 77.4\\
BEiT-B           & 63.8 & 76.6 & 83.7 & 80.0 & 84.3 & 77.7\\
BEiT-L         & 66.4 & 77.9 & \textbf{85.1} & 82.9 & 81.3 & 78.7\\
OneFormer     & 61.0 & 75.7 & 83.9 & \textbf{84.1} & \textbf{91.5 }& 79.2\\
EoMT          & \textbf{70.1} &\textbf{ 80.0} & 84.6 & 82.0 & 87.3 & \textbf{80.8}\\
\bottomrule
\end{tabular}
\caption{Class-wise IoU and mIoU (\%) for each model on \textit{EIDSeg}.  Best scores are \textbf{bold}.}
\label{tab:miou_class_iou}
\end{table*}
\textbf{Dataset Preparation:} We follow the data splitting strategy adopted in the xBD paper~\cite{Gupta2019xBD} and divide the EIDSeg dataset into 80\% for training, 10\% for validation and 10\% for testing.

\textbf{Models and Training:} We evaluated eight state-of-the-art semantic segmentation models: DeepLabV3+(ResNet-101)~\cite{Chen2018DeepLabV3Plus}, SegFormer (B5)~\cite{Xie2021SegFormer}, Mask2Former (Swin-Base and Swin-L)~\cite{Cheng2022Mask2Former}, BEiT (ViT-B and ViT-L)~\cite{Bao2022BEiT}, OneFormer (Swin-L)~\cite{Jain2023OneFormer}, and EoMT (ViT-L)~\cite{Kerssies2025EoMT}. All models are initialized with Cityscapes-pre-trained weights except for BEiT, which uses the ADE20K-pre-trained checkpoint due to the absence of a Cityscapes version.

To accommodate the architectural constraints of each model, we adopt model-specific input resolutions: 640$\times$640 for BEiT; 1024$\times$1024 for EoMT; and 512$\times$512 for other models. During training, images are augmented with random horizontal flips, color jittering, and random cropping. All models are trained using the AdamW optimizer with a learning rate of $1\times10^{-5}$, $\beta_1 = 0.9$, $\beta_2 = 0.999$, and a weight decay of 0.01. Pixel-wise cross-entropy loss is used for supervision. The learning rate is scheduled using the \texttt{ReduceLROnPlateau} strategy with patience of 10 epochs and a reduction factor of 0.5. Training runs for up to 100 epochs, with early stopping triggered after 30 epochs of no improvement on the validation set. All experiments are run on a single NVIDIA A100-80GB GPU and a batch size of 4. The same training protocol is used for all models to ensure a fair and consistent comparison.

\textbf{Evaluation:} We use standard evaluation metrics that include mIoU, frequency-weighted intersection over union (FWIoU), pixel accuracy (PA), floating point operations (FLOPs), and the number of model parameters. FLOPs are computed using the \texttt{fvcore} library \footnote{\url{https://github.com/facebookresearch/fvcore}}, and reported in terms of GFLOPs (i.e., FLOPs $\times\ 10^9$).

\subsection{Benchmark Analysis}
As discussed in Related Work, the DSS dataset is the only post-earthquake social media dataset for semantic segmentation. The original DSS dataset reported only one baseline: SegFormer‑B5 trained for three-class segmentation, which achieved an mIoU of 72\%~\cite{Zhang2025PixelsDamageSeverity}. In contrast, Table~\ref{tab:seg_results} presents a comprehensive evaluation of eight modern architectures on the proposed \textit{EIDSeg} benchmark, which adopts a more fine-grained five-class labeling. Under these conditions, SegFormer‑B5 attains an mIoU of \textbf{74.4\%}, yielding a \textbf{2.4\%} improvement over the DSS result while addressing a more detailed five-class labeling that includes two additional categories. This gain highlights the advantages of our expanded image corpus and refined annotation strategy, which collectively improve class balance and reduce labeling noise.

EoMT achieves the highest overall performance, with a FWIoU of \textbf{80.9\%} and pixel accuracy of \textbf{90.3\%}. Its superior results can be attributed to training in larger input resolutions (\(1024 \times 1024\)) that preserve spatial detail, as well as recent architectural advances in transformer design. Other transformer-based models, including Mask2Former, BEiT, and OneFormer, exhibit strong performance, demonstrating that EIDSeg effectively supports generalization across diverse model families.

Table~\ref{tab:miou_class_iou} presents the class-wise IoU for each model. Among all classes, \textit{Undamaged Building} and \textit{Damaged Building} consistently exhibit lower IoU scores. This can be attributed to two main factors. First, during the annotation process for EIDSeg, the annotators followed a conservative guideline: if a building appeared structurally intact but exhibited visible cracks not attributable to wear and tear, it was labeled as \textit{Damaged Building} due to lack of information about its interior condition. Although this labeling strategy supports cautious decision-making in post-disaster analysis, it may lead to confusion for models when encountering visually similar undamaged structures. Second, since the dataset is collected from social media, buildings in the background often appear blurry or lack sufficient detail, making it difficult to assess structural damage. This reflects an inherent trade-off in the use of social media imagery. Although it enables access to timely data from disaster-stricken areas where professionals may not be present, it also limits control over image quality and resolution. Visual examples of misclassified images and the leave-one-event-out results are presented in Appendix B and Appendix D, respectively.

\section{Conclusion}
 In this work, we introduce EIDSeg, the largest pixel-level semantic segmentation dataset for post-earthquake damage analysis based on social media images. The dataset comprises 3,266 images collected from nine major earthquakes that occurred between 2008 and 2023, covering a diverse range of real-world disaster scenarios. Annotations span five semantic classes: Undamaged Building, Damaged Building, Destroyed Building, Undamaged Road, and Damaged Road, allowing a fine-grained evaluation of damage to built infrastructure. Through a carefully designed annotation guideline and expert-led quality control, EIDSeg provides a reliable foundation for training vision models capable of automating post-earthquake damage assessment using ground-view imagery.

\section{Acknowledgment}
The research described herein was supported in part by the School of Computing Instruction and the Elizabeth and Bill Higginbotham Professorship at Georgia Tech. This support is gratefully acknowledged. Special thanks are due to Mohammad Taher for helping with the annotation. Support for the undergraduate students participating in the project was provided by the US National Science Foundation through the Geotechnical Engineering Program under Grant No. CMMI-1826118. Any opinions, findings, and conclusions or recommendations expressed in this material are those of the authors and do not necessarily reflect the views of NSF.

% \section{Acknowledgment}

\bibliography{aaai2026}

\appendix
\onecolumn       
\definecolor{undesignatedColor}{HTML}{BF00FF}

\section{Appendix}
\begin{figure*}[th!]
    \centering
    \includegraphics[width=0.9\linewidth]{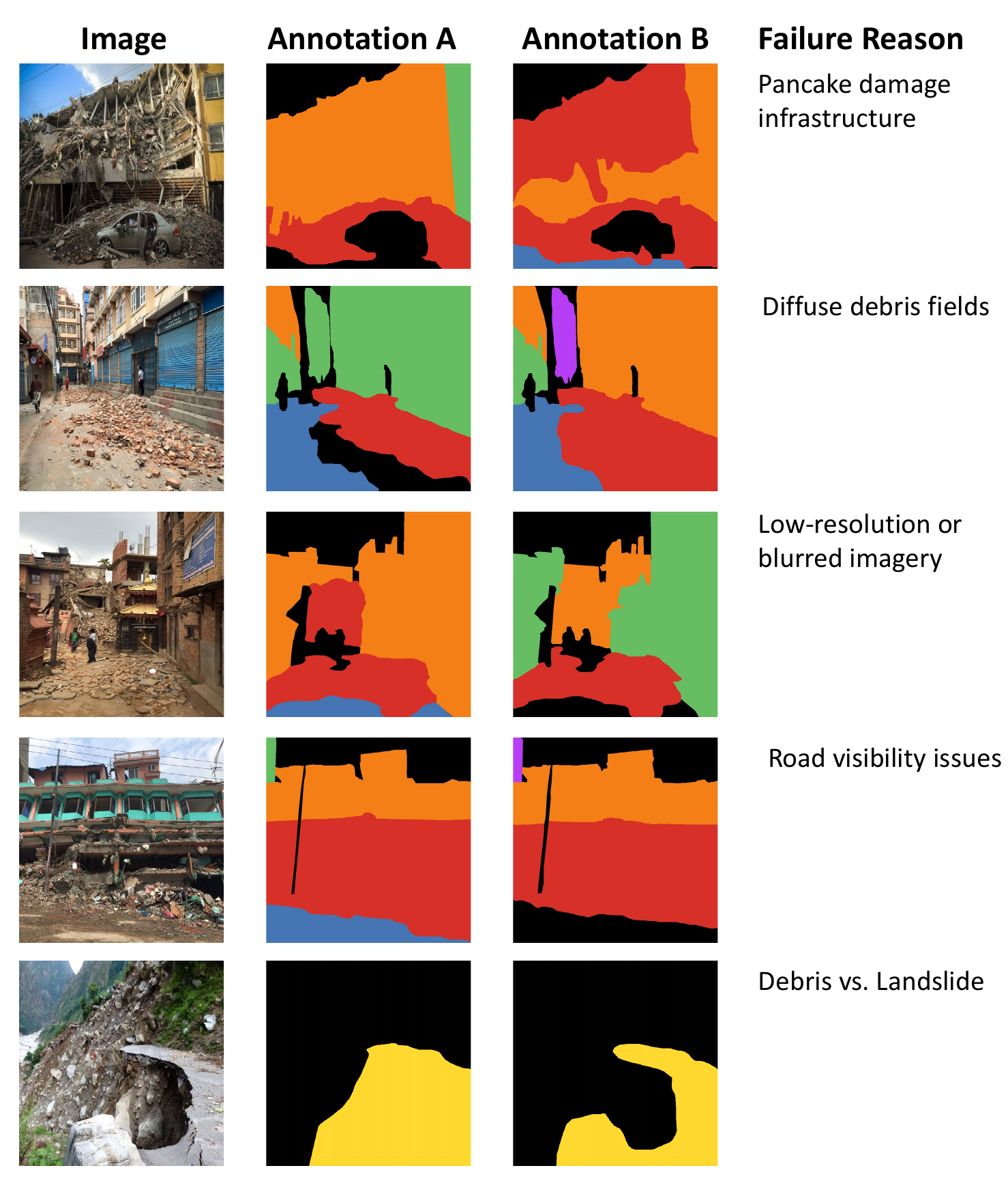}
    \caption{Examples of annotator misalignment observed during the annotation process. Annotation~A and Annotation~B show labels produced by two annotators within the same group. The failure reasons are summarized in the rightmost column, with additional details provided in Appendix~A. Class colors correspond to the semantic segmentation labels: 
\textcolor{green!50!black}{Undamaged Building}, 
\textcolor{orange}{Damaged Building}, 
\textcolor{red!70!black}{Debris}, 
\textcolor{blue}{Undamaged Road}, 
\textcolor{yellow!90!black}{Damaged Road}, 
and the \textcolor{undesignatedColor}{Undesignated} category.}
    \label{fig:annotators_disagreement}
\end{figure*}

\begin{figure*}[t!]
    \centering
    \includegraphics[width=1.0\linewidth]{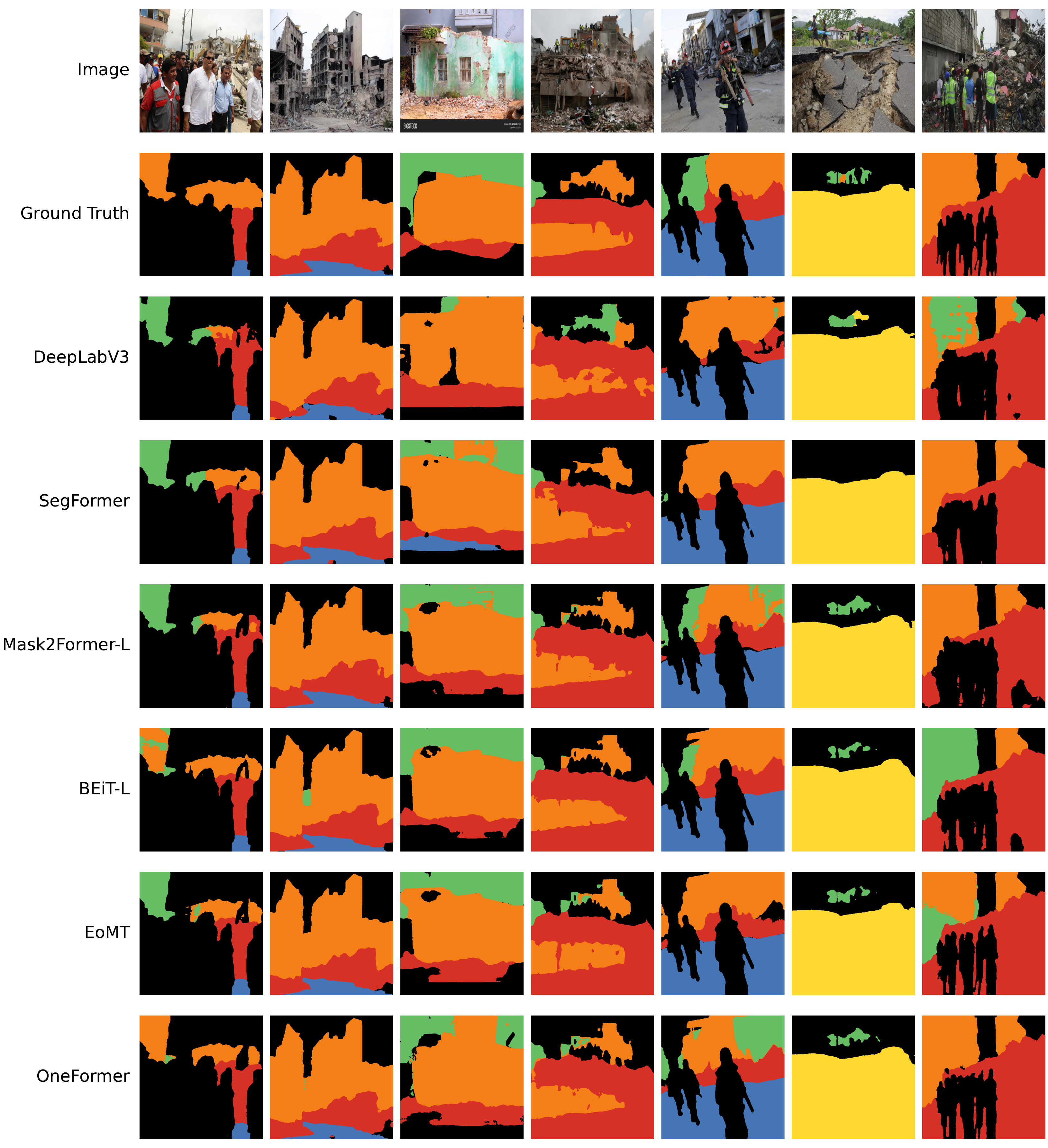}
    \caption{Examples of misclassifications across different models.Class colors correspond to the semantic segmentation labels: 
\textcolor{green!50!black}{Undamaged Building}, 
\textcolor{orange}{Damaged Building}, 
\textcolor{red!70!black}{Debris}, 
\textcolor{blue}{Undamaged Road}, 
and \textcolor{yellow!90!black}{Damaged Road} category.}
    \label{fig:misclassified_examples}
\end{figure*}

\subsection{A. Annotators Misalignment Examples}\label{misalignment}

Figure~\ref{fig:annotators_disagreement} shows representative examples of annotator disagreement observed during the EIDSeg labeling process. These cases summarize the most common sources of ambiguity encountered in post–earthquake scenes.

\begin{itemize}
\item \textbf{Pancake damage infrastructure (R1).} The image shows a partially collapsed building where upper floors have failed while lower floors remain standing. Annotation~A labels the entire structure as a damaged building, while Annotation~B assigns the collapsed portion to \emph{Debris}. Following DSS annotation practice~\cite{Zhang2025PixelsDamageSeverity}, we segment pancake collapsed components as \emph{Debris} rather than \emph{Damaged Building}.

\item \textbf{Diffuse debris fields (R2).} Earthquake rubble often appears as scattered and irregular fragments without a clear boundary. Annotators differed on whether to segment only the main debris cluster or to include all surrounding fragments.

\item \textbf{Low-resolution or blurred imagery (R3).} Blurred buildings and roads reduce the visibility of structural details such as facade cracks or damaged surfaces. As a result, some annotators under-labeled these regions.

\item \textbf{Road visibility issues (R4).} Roads partially obscured by dust, rubble, or shadows were inconsistently recognized. In several cases, the road surface was omitted even though it remains part of the infrastructure class.

\item \textbf{Debris vs. landslide material (R5).} In mountain environments, it is difficult to distinguish earthquake-generated debris from natural landslide material. In this example, Annotator~A labels the lower region as debris, while Annotator~B interprets it as geologic material. The final annotation is shown in Figure~\ref{fig:eidseg_mask}.

\end{itemize}

\subsection{B. Visualization of Misclassified Examples}

Figure~\ref{fig:misclassified_examples} presents a representative collection of misclassified cases across all evaluated models. Several consistent error patterns are apparent. First, blurred or low-resolution regions substantially reduce segmentation accuracy. In the sixth column, for example, every model fails to identify the damaged building in the background due to the loss of fine structural detail. Second, fragmented structural predictions occur frequently. Rather than segmenting a building as a continuous object, several models break it into multiple disconnected regions. This reflects a heavy dependence on localized texture cues instead of holistic shape information, a limitation that becomes especially pronounced in social-media photographs characterized by distortion, shadows, clutter, and occlusion.

Across disaster contexts, these qualitative patterns illustrate the inherent challenges of post-event street-level imagery, where fluctuating image quality, heterogeneous construction materials, and complex debris fields introduce significant visual ambiguity. The examples further show that even state-of-the-art segmentation networks exhibit systematic weaknesses in delineating damaged urban infrastructure. These observations underscore the need for stronger structural priors, enhanced robustness to image degradation, and improved handling of occlusions in future segmentation approaches.

\subsection{C. Histogram of the Images before Data Cleaning}

\begin{figure*}[htbp]
    \centering
    \includegraphics[width=0.85\linewidth]{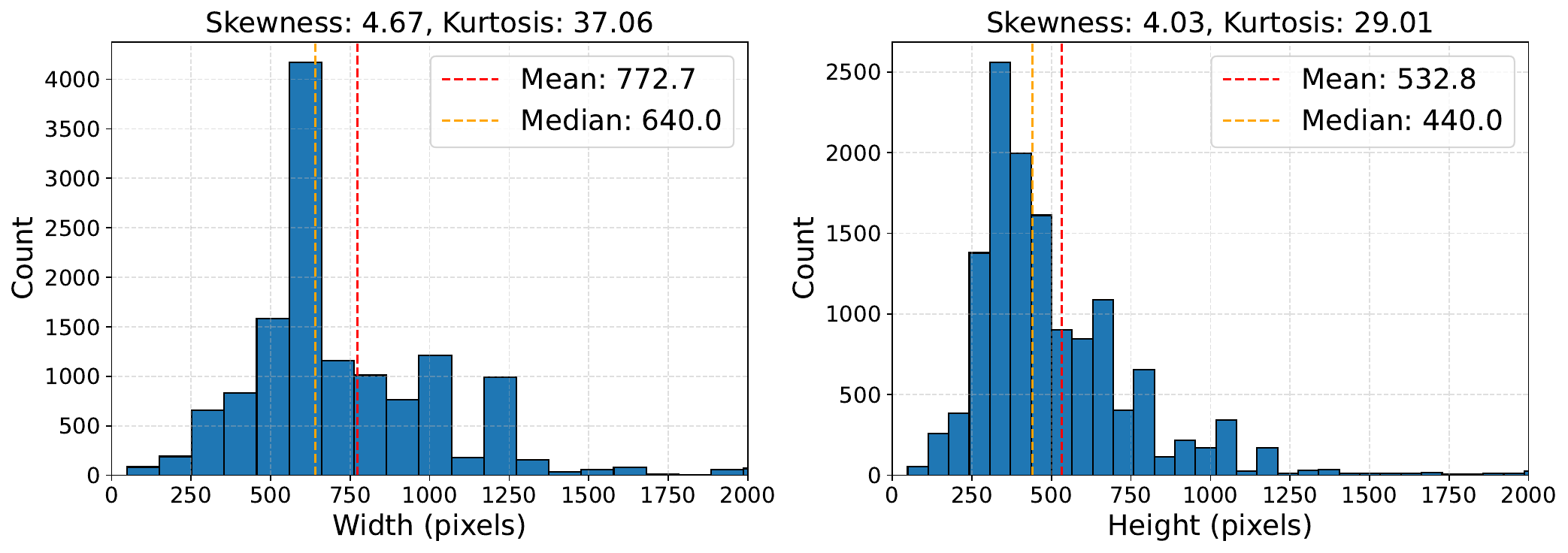}
    \caption{Distribution of image widths and heights in the EID dataset~\cite{Huang2025Fidelity} 
    The red dashed lines indicate the mean values, and the orange dashed lines indicate the medians. 
    Skewness and kurtosis are reported above each histogram, highlighting the strong positive skew and heavy-tailed nature of the resolution distribution. The $x$-axis is truncated at 2000 pixels for clarity, since images with larger dimensions are rare.}
    \label{fig:eid_histogram}
\end{figure*}

For the DSS dataset, which has already undergone expert curation, the overall image quality is high. To maintain consistency with the EID dataset, we further removed low-resolution samples by filtering images according to the resolution distribution observed in the EID. Figure~\ref{fig:eid_histogram} illustrates the distribution of image resolutions in the EID dataset. The resolution threshold described in the main paper was derived from the pixel-size distribution across all samples. Applying this criterion led to the exclusion of approximately 3,050 images (21.6\%), the majority of which were non–disaster related. Among the 58 cases manually reconsidered by experts, only 2 remained after the separate group annotation process, further confirming that low-resolution imagery is unsuitable for reliable segmentation.

\subsection{D. Leave-One-Event-Out Results }

The leave-one-disaster-out evaluation with the EoMT model requires identifying the source disaster for every image. Because EID~\cite{Huang2025Fidelity} and DSS~\cite{Zhang2025PixelsDamageSeverity} aggregate material from multiple online sources without full provenance, the analysis includes only images with verifiable origins. This restriction ensures that every split corresponds uniquely to one event.

Table~\ref{tab:disaster_performance_sorted} lists the results for all nine disasters. Performance varies across events, with Chile, Ecuador, and Turkey reaching the highest FWIoU and PA values. Haiti and Iraq--Iran show lower scores, reflecting greater visual variation and less transferable damage patterns. Averaged across all disasters, the model reaches 80.76\% FWIoU and 88.64\% PA, indicating consistent cross-event behavior under the leave-one-out protocol.

\begin{table}[!h]\
\centering
\begin{tabular}{lcc}
\hline
\textbf{Disaster} & \textbf{FWIoU (\%)} & \textbf{PA (\%)} \\
\hline
Haiti              & 74.46 & 85.16 \\
Iraq--Iran         & 74.86 & 84.80 \\
Morocco            & 77.86 & 87.58 \\
Mexico             & 81.85 & 90.00 \\
Wenchuan           & 82.04 & 89.79 \\
Nepal              & 82.48 & 90.07 \\
Turkey             & 83.07 & 90.62 \\
Ecuador            & 84.21 & 91.23 \\
Chile              & 86.05 & 92.45 \\
\hline
\textbf{Average}   & \textbf{80.76} & \textbf{88.64} \\
\hline
\end{tabular}
\caption{Semantic segmentation performance of EoMT across different disasters (sorted by FWIoU ascending).}
\label{tab:disaster_performance_sorted}
\end{table}

\subsection{E. Computational Performance Analysis}

\begin{table*}[ht]
\centering
%\scriptsize
%\setlength{\tabcolsep}{11pt}
\begin{tabular}{lcccccc}
\hline
\textbf{Model} &
\textbf{A100 Lat.} &
\textbf{A100 Mem.} &
\textbf{A100 Thr.} &
\textbf{CPU Lat.} &
\textbf{CPU Mem.} &
\textbf{CPU Thr.} \\
 & (ms) & (MB) & (img/s) & (ms) & (MB) & (img/s) \\
\hline
DeepLabV3+    & 76.4  & 442/496   & 12   & 327   & 944  & 3.06 \\
SegFormer     & 37.6  & 1669/1704 & 26.4 & 881   & 1164 & 1.13 \\
Mask2Former-S         & 37.2  & 704/888   & 26.9 & 750   & 2251 & 1.33 \\
Mask2Former-L         & 37.6  & 1198/1446 & 26.6 & 1548  & 2412 & 0.65 \\
BEiT-B        & 60.3  & 13192/13298 & 16.6 & 3417 & 2088 & 0.29 \\
BEiT-L        & 206.8 & 23939/24034 & 4.8  & 7683 & 3573 & 0.13 \\
OneFormer     & 43.5  & 1528/1928 & 23   & 1723  & 2517 & 0.58 \\
EoMT          & 124.9 & 5814/6138 & 8    & 8494  & 2405 & 0.12 \\
\hline
\end{tabular}
\caption{Performance comparison on GPU (A100,batch=1) and CPU (ONXX runtime, single thread). 
\textbf{Lat.}: single-image latency (ms). 
\textbf{Mem.}: GPU (reserved/allocated) memory; CPU peak allocated memory from ONNX Runtime. 
\textbf{Thr.}: throughput (img/s). }

\label{tab:model_perf}
\end{table*}

Table~\ref{tab:model_perf} summarizes latency, throughput, and memory usage for all models on an A100 GPU and an ONNX Runtime CPU proxy. On the A100, the query-based architectures (Mask2Former-S/L) provide the most favorable balance of speed and memory, sustaining $\sim$37\,ms latency and $\sim$27\,img/s throughput. SegFormer achieves similar throughput but reserves substantially more memory, while DeepLabV3+ remains the most memory-efficient at the cost of lower throughput. Transformer-heavy designs (BEiT-B/L and EoMT) exhibit significantly higher latency and memory demand, with BEiT-L exceeding 24\,GB even at batch size~1.

The CPU results further accentuate these differences. DeepLabV3+ delivers the highest throughput (3.06\,img/s) and the lowest latency among all models, reflecting the efficiency of convolutional operators on commodity hardware. Mask2Former-S/L and SegFormer form a middle tier (1.1--1.3\,img/s), whereas BEiT-B/L and EoMT fall below 0.3\,img/s.

Overall, Mask2Former offers the best trade-off on high-end GPUs, while DeepLabV3+ is the most practical choice for CPU- and edge-class deployment.

\end{document}